\pdfoutput=1 
\documentclass{article}

\newif\ifarxiv
\arxivtrue

\ifarxiv
  \usepackage[preprint]{neurips_2026}
\else
  \usepackage{neurips_2026}
\fi

\usepackage[utf8]{inputenc} 
\usepackage[T1]{fontenc}    
\usepackage{hyperref}       
\usepackage{url}            
\usepackage{booktabs}       
\usepackage{amsfonts}       
\usepackage{microtype}      
\usepackage{xcolor}         
\usepackage{amsmath}
\usepackage{amssymb}
\usepackage{array}
\usepackage{enumitem}
\usepackage{graphicx}

\graphicspath{{figures/}}
\newcolumntype{L}[1]{>{\raggedright\arraybackslash}p{#1}}

\title{Designing Reliable LLM-as-a-Judge Measurement Systems for Multi-Turn Business Agents}

\author{%
  Kaiwen Luo \\
  Meta \\
  \texttt{kaiwenluo@meta.com} \\
  \And
  Ming Gao \\
  Meta \\
  \texttt{gaoming@meta.com} \\
}

\begin{document}

\maketitle

\begin{abstract}
Many LLM-as-a-judge evaluations score fixed outputs under a fixed task definition. Production
multi-turn business agents instead require a maintained measurement system: correctness depends on
business-specific facts and procedures, outcomes emerge across turns, and failures must be attributed to
either agent capability or missing business knowledge before they are actionable.
We present an integrated methodology spanning evaluation specification, modular LLM judges,
intent-preserving user simulation, and human-in-the-loop governance. The specification defines
conversation-level end states and actionable failure ownership. Atomic judges share versioned evidence and
feed an explicit aggregation graph. The simulator is released only after task-preservation and stability
checks. Independent human audits estimate measurement fidelity, renew tiered reference sets, and route
disagreements to label correction, guideline revision, or judge improvement.
Production studies show that system-level fidelity improved across repeated audits, that human reviewers
and automated judges improved together under the shared feedback loop, and that their combined workflow
had the strongest descriptive performance in both reported task-completion settings. Because the studies are observational
and the human reference itself required revision, these findings demonstrate operational usefulness rather
than causal or universal superiority. The contribution is a practical framework for making multi-turn
agent measurement reliable, actionable, and maintainable as the evaluated system and its evidence evolve.
\end{abstract}

\section{Introduction}
\label{sec:intro}

LLM-as-a-judge is now an established approach for evaluating model outputs. Strong models can rank responses, apply structured rubrics, and approximate human preferences \citep{zheng2023judging,liu2023geval,kim2023prometheus}, while meta-evaluation benchmarks increasingly characterise judge reliability \citep{tan2024judgebench,thakur2024judgingjudges}. Most of this work, however, evaluates fixed outputs under a fixed task definition. Multi-turn benchmarks measure conversational capability \citep{kwan2024mteval,bai2024mtbench101,guan2025multiturnsurvey}, but how to build and maintain a continuously operated LLM-as-a-judge measurement system for multi-turn trajectories remains less well understood. The object of evaluation is no longer one response: task completion, state consistency, and failure attribution emerge across the interaction.

Business agents make this problem harder still. An agent answers from one merchant's catalogue, FAQ, price list, and written procedures, so correctness is instance-specific rather than defined by one golden answer. Whether ``we ship to that address in three days'' is correct depends on a fact the evaluator may not have; whether the agent should request an order number before an email address depends on a merchant-authored procedure. When the interaction fails, the operational question is not merely whether the response was poor, but whether the model failed or the business never supplied the required data.

These conditions turn LLM-as-a-judge from a prompt-design problem into a measurement-system problem. The system must define a conversation-level construct, ground judgments in per-business evidence, separate failure owners, and remain calibrated as guidelines, products, and traffic change. A standard judge over a reference answer, preference pair, or cross-instance rubric does not provide these properties by itself.

Direct user feedback is insufficient as a primary measurement source. Production evidence shows that
explicit feedback can be both sparse and concentrated among a small subset of users
\citep{maharaj2024enterprise}, making it unreliable for traffic-level quality measurement on its own.

Human review is necessary but is not, by itself, a scalable or automatically reliable measurement system.
Its fidelity must also be measured against an independently audited reference; the empirical baselines are reported in
Section~\ref{sec:validation}.

\subsection{Proposed measurement-system methodology}
\label{sec:1-1}

\textbf{We propose a methodology for designing and operating LLM-as-a-judge measurement systems for multi-turn business agents.}

\begin{center}
\footnotesize
\setlength{\tabcolsep}{3pt}
\renewcommand{\arraystretch}{1.08}
\begin{tabular}{@{}L{3.1cm} L{4.2cm} L{5.4cm}@{}}
\toprule
Component & What it produces & The design rule it must satisfy \\
\midrule
\textbf{Evaluation specification} (Section~\ref{sec:spec}) & constructs, metrics, labeling guidelines, and the unit of analysis & define an observable and actionable measurement target before implementing judges \\
\textbf{LLM-as-a-judge architecture} (Section~\ref{sec:judgment}) & scoped judges, shared context, and an explicit aggregation graph & implement the specification through modular judgments and explicit composition rules \\
\textbf{LLM-as-a-user simulation} (Section~\ref{sec:simulation}) & adaptive pre-launch trajectories and simulator-fidelity measurements & preserve the intended user task and validate the simulator before judging the agent \\
\textbf{Human-in-the-loop evaluation and improvement} (Section~\ref{sec:human}) & fidelity audits, renewable reference labels, and correction channels & evaluate the system and route feedback to labels, specifications, and judges \\
\bottomrule
\end{tabular}
\end{center}

The methodological contribution lies in the interfaces between these components. The judge prompt is the executable form of the specification, so guideline changes require measurement releases. Judge--human disagreements update the specification as well as individual labels. The same judge implementation runs offline and online, allowing observed discrepancies to be attributed to evaluation data rather than evaluator differences. The simulated user consumes versioned intents, answers, and user facts; its intent consistency, persona consistency, comprehensibility, and stability are evaluated separately from agent quality. Human audit renews the reference set used to validate every component. Measurement fidelity, quantified against an audited human reference, provides the system-level control signal for these updates.

We do not claim novelty in the individual techniques. Decomposing a judge into per-dimension questions is established practice \citep{ye2023flask,lee2024checkeval,li2024dnaeval,liu2024hdeval,cook2024ticking,hashemi2024llmrubric,wei2025rocketeval,zhang2026rethinkingatomic,cho2026bineval}. Conversation-level evaluation is established \citep{mehri2020fed,zhou2023sotopia,arora2025healthbench,fan2026shoppingreasoning}. LLM user simulation is established \citep{yao2024taubench,barres2025tau2bench,arcadinho2024almita}. Human-in-the-loop judge alignment is established \citep{shankar2024validators,ashktorab2025evalassist}. Our contribution is a system-level methodology that connects specification, judge development, intent-preserving LLM-as-a-user simulation, and human-in-the-loop evaluation and improvement. Section~\ref{sec:related} positions this claim against the closest work.

\subsection{Contributions}
\label{sec:1-2}

\begin{enumerate}[leftmargin=1.6em, itemsep=0.4ex, topsep=0.4ex]
\item \textbf{A problem formulation for LLM-based measurement (Section~\ref{sec:setting}).} Four properties of multi-turn business agents --- per-business ground truth, procedure adherence, model-gap-versus-data-gap attribution, and conversation-scoped outcomes --- that jointly make generic judge recipes inapplicable, with the metric taxonomy we adopted in response.
\item \textbf{An end-to-end measurement-system methodology (Sections~\ref{sec:spec}--\ref{sec:human}).} A lifecycle connecting evaluation specification, LLM-as-a-judge development, intent-preserving LLM-as-a-user simulation, and human-in-the-loop evaluation and improvement. The contribution lies in the complete lifecycle and its cross-component constraints rather than any individual judge technique or fidelity metric.
\item \textbf{Empirical studies (Section~\ref{sec:validation}).} Production studies examine longitudinal
improvement in measurement fidelity and complementarity between human raters and judges.
\end{enumerate}

\section{Problem formulation}
\label{sec:setting}

We formulate the requirements that an LLM-as-a-judge measurement system must satisfy in this setting. Each stage of the methodology in Sections~\ref{sec:spec}--\ref{sec:human} addresses one or more of these requirements.

\paragraph{P1 --- There is no single golden answer, and correctness is business-specific.}
Many semantically equivalent answers can be valid, while the same factual statement can be correct for one business and a hallucination for another because the underlying catalogue or policy differs. Reference-based metrics and cross-instance rubrics are therefore insufficient on their own.

\paragraph{P2 --- Correctness includes procedure adherence, which is written by a third party.}
Businesses author their own operating procedures, including ordered information-gathering and authentication rules, and following them is part of quality. Judging this requires applying a business-authored document per conversation, and it is a frequent source of human--judge disagreement.

\paragraph{P3 --- A failure must be attributed before it is actionable.}
The organisation operating the agent can fix a model; it cannot fix a fact the merchant never supplied --- it can only ask for it. So the topline taxonomy separates \textbf{model-capability gaps} (the agent could have answered and did not) from \textbf{knowledge-coverage gaps} (the required information was absent from the business's data). These have different owners and different remediations, and an evaluation that reports only ``bad conversation'' is not usable by either. This attribution requirement is what forces an explicit aggregation layer (Section~\ref{sec:judgment}) --- and it is also what makes the topline the hardest thing in the taxonomy to measure, because exclusive attribution is a conjunction with a negation.

\paragraph{P4 --- The defects that matter are conversation-scoped.}
The failures that actually damage a multi-turn interaction are cross-turn state defects: the agent re-asks for information the customer already gave, contradicts an earlier commitment, or fails to notice that the customer is repeating themselves because they are not being helped. None of these is visible in any single turn, to a human or to a judge --- the same blind spot reported independently on production transaction agents \citep{zhang2026catchingonefive}. A turn-level rubric also asks raters a counterfactual question (``could this response have harmed the interaction?''), motivating conversation-level labels whose empirical reliability is reported in Section~\ref{sec:validation}.

\section{Related work}
\label{sec:related}

The components of our methodology are individually well established. The relevant question is how they
are coupled and maintained for multi-turn business-agent measurement.

\subsection{Prior work by measurement-system function}
\label{sec:3-1}

Strong models can approximate human preferences, but benchmark agreement is not equivalent to validity
\citep{zheng2023judging,kim2023prometheus,tan2024judgebench,li2024generation}. Top-venue work on
fine-grained rubrics, checklists, and decomposition makes judgments more inspectable
\citep{ye2023flask,lee2024checkeval,li2024dnaeval,wei2025rocketeval,viswanathan2025checklists}, while
Autorubric makes criterion design, weighting, aggregation, abstention, and calibration explicit
configuration choices \citep{rao2026autorubric}. We therefore claim novelty neither for decomposition nor
for aggregation structure. Established findings on position bias and statistical calibration instead
motivate validating each dimension and the resulting composite against human labels
\citep{shi2024judgingposition,lee2025correctly}.

Multi-turn evaluation is also mature. MINT and MT-Bench-101 test capabilities that emerge across turns
\citep{wang2023mint,bai2024mtbench101}, while CRMArena evaluates agents on realistic business tasks
\citep{huang2024crmarena}. Simulated-user benchmarks such as $\tau$-bench extend evaluation to adaptive
agent--user interaction \citep{yao2024taubench}. Our contribution is not conversation-level evaluation or
user simulation alone, but their inclusion in a measurement lifecycle where task preservation,
information boundaries, and simulator stability are release criteria.

\subsection{From judge methods to measurement systems}
\label{sec:3-2}

Human labels are themselves part of the instrument: disagreement can be substantive, while guideline
design can create avoidable error \citep{plank2022labelvariation,parmar2022dontblame}. The most comparable
calibration protocol gives humans and the automated evaluator the same instructions and questions
\citep{chiang2023can}, and label-error research shows why automated disagreements still require human
adjudication \citep{northcutt2021pervasive}. Production precedents include end-to-end taxonomy, labeling,
and monitoring for safety classifiers \citep{markov2022moderation}, deployed customer-service evaluation tied to
business outcomes \citep{xu2024ragkg}, and an
evaluation-driven framework for customer-support agents with human-in-the-loop judge development and
online validation \citep{gupta2026nubank}.

Accordingly, we make no claim to the first decomposed judge, evaluation graph, simulator, or continuously
maintained evaluator. The contribution is the integration required for multi-turn business trajectories:
per-business facts and procedures, conversation-level outcomes, explicit
model-gap-versus-knowledge-gap attribution, simulator validation, composite-node fidelity, and recurring
human audit are governed as one measurement system.

\section{Evaluation specification}
\label{sec:spec}

Evaluation specification determines what the system should measure before any judge is built. For multi-turn business agents, the specification must resolve four questions: what object is evaluated, what counts as success, how failures are assigned to an actionable owner, and what business-specific evidence supports the verdict. \textbf{Design principle: define and validate the measurement target before optimizing its automated implementation.}

\subsection{From evaluation requirements to metric definition}
\label{sec:spec-metrics}

The properties in Section~\ref{sec:setting} imply that a generic response-quality score is insufficient. Cross-turn failures require the \textbf{conversation} as the unit of analysis; operational remediation requires separating an agent-capability failure from missing business knowledge; and factual correctness and procedure adherence must be grounded in each business's catalogue, FAQ, business information, and operating procedures.

The resulting specification asks two questions for every conversation: \textbf{were all valid user tasks completed, and if not, what prevented completion?} Invalid and unfinished interactions are handled separately so that an unresolvable or truncated conversation is not automatically counted as an agent failure. Each conversation is assigned to one of seven mutually exclusive end states: \emph{full resolution, partial resolution, invalid conversation handled well, unfinished conversation handled well, model-capability gap only, knowledge-coverage gap only,} or \emph{both gaps}. Two topline rates follow:

\begin{itemize}[leftmargin=1.6em, itemsep=0.4ex, topsep=0.4ex]
\item \textbf{model success rate} = 1 $-$ rate of conversations with a model-capability gap
\item \textbf{knowledge coverage rate} = 1 $-$ rate of conversations with a knowledge-coverage gap
\end{itemize}

Below these rates, the specification retains progressively finer root causes for diagnosis: L1 records success or failure, L2 records the end state, and L3--L5 identify dimensions such as missing data, hallucination, action accuracy, and procedure following. This hierarchy separates the reported metric from the evidence used to explain it and later determines the aggregation graph implemented by the judge system.

\subsection{Reference-label reliability and measurement fidelity}
\label{sec:reference-fidelity}

Nothing downstream is interpretable until the human baseline is known. The relevant question is not
whether a judge can ``beat a human,'' but whether either labeling mechanism produces a faithful measurement
against an independently reviewed reference. The observed baselines are reported in
Section~\ref{sec:validation}.

\paragraph{Agreement is not fidelity.}
Raw rater consistency can fall while audited label quality improves when raters move from near-unanimous negative labeling to detecting more positives. Krippendorff's $\alpha$ corrects for chance agreement but remains prevalence-sensitive and does not measure correctness. Agreement metrics are therefore retained as reliability diagnostics, while fidelity is measured against an independently audited reference.

\paragraph{Net detection rate is the system-level fidelity metric.}
Let $p$ be the true prevalence of the target condition, $\mathrm{TPR}$ and $\mathrm{FPR}$ the labeling system's true- and false-positive rates, and $y$ the observed positive rate. The positive class is metric-specific: it denotes the presence of a defect for gap metrics and successful completion or resolution for success metrics. Then

\begin{equation}
\mathbb{E}[y] = \mathrm{FPR} + (\mathrm{TPR}-\mathrm{FPR})p = \mathrm{FPR} + \mathrm{NDR}\,p,
\end{equation}

Following Youden's original diagnostic-test index \citep{youden1950index}, we define \textbf{net detection rate} as $\mathrm{NDR}=\mathrm{TPR}-\mathrm{FPR}=\text{sensitivity}+\text{specificity}-1$ (Youden's $J$). NDR ranges from $-1$ to 1: 1 is perfect measurement, 0 carries no information about the true label beyond chance, and a negative value reverses the direction of the signal.

NDR is the appropriate control objective for three reasons. First, if TPR and FPR remain stable across the populations being compared, metric movement is attenuated as $\mathbb{E}[\text{measured }\Delta]=\mathrm{NDR}\cdot\text{true }\Delta$; maximizing NDR preserves more of the underlying product movement. Second, correcting prevalence requires division by NDR, so the variance of the corrected estimate scales with $1/\mathrm{NDR}^2$; higher NDR therefore supports tighter confidence intervals and smaller detectable effects. Third, unlike precision, NDR is defined by TPR and FPR and does not change mechanically when defect prevalence changes.

Precision, recall, F1, and inter-rater agreement remain useful guardrails and diagnostic metrics, but they do not replace NDR as the topline measure of measurement fidelity. The attenuation interpretation also depends on stable error rates, so NDR must be re-estimated after changes to the product, traffic, guideline, reference process, judge prompt, or foundation model.

\section{LLM-as-a-judge architecture}
\label{sec:judgment}

This stage translates the evaluation specification into executable judges, shared inputs, and aggregation logic. \textbf{Design principle: implement the specification through modular judges, governed inputs, and an explicit aggregation graph.}

The system uses per-dimension atomic judges and an explicit aggregation layer rather than a single root-cause judge. Section~\ref{sec:related} establishes that decomposition is settled prior work; our focus is its integration into a versioned measurement architecture.

Three properties of the construction matter for the rest of the section. \textbf{(i) Independent judgments} --- each judge is a standalone unit with a uniform input/output contract emitting a binary verdict plus free-text reasoning; unlike the human labeling process, in which checkers are prompted sequentially, the judges assess all dimensions in parallel. \textbf{(ii) The aggregation layer is separate and separately versioned} --- metric and attribution definitions are changed by editing the graph rather than by silently changing judge behaviour. \textbf{(iii) A single versioned context store} --- all grounding evidence a judge sees resolves through one store rather than per-judge lookups, eliminating the class of disagreement in which two judges score the same conversation against different facts. That last one is a direct response to P1 (Section~\ref{sec:setting}): with business-specific ground truth, \emph{which facts was the judge holding?} is part of the verdict.

The architecture is implemented as a versioned registry shared across agents, with a scoped subset invoked
for each conversation. Composite dimensions are themselves decomposed: the missing-data detector first summarises the
conversation into discrete user asks and then judges each independently, turning a multi-class problem
into a set of binary ones and keeping greeting turns out of the judged text.

\begin{figure}[t]
\centering
\includegraphics[width=\textwidth]{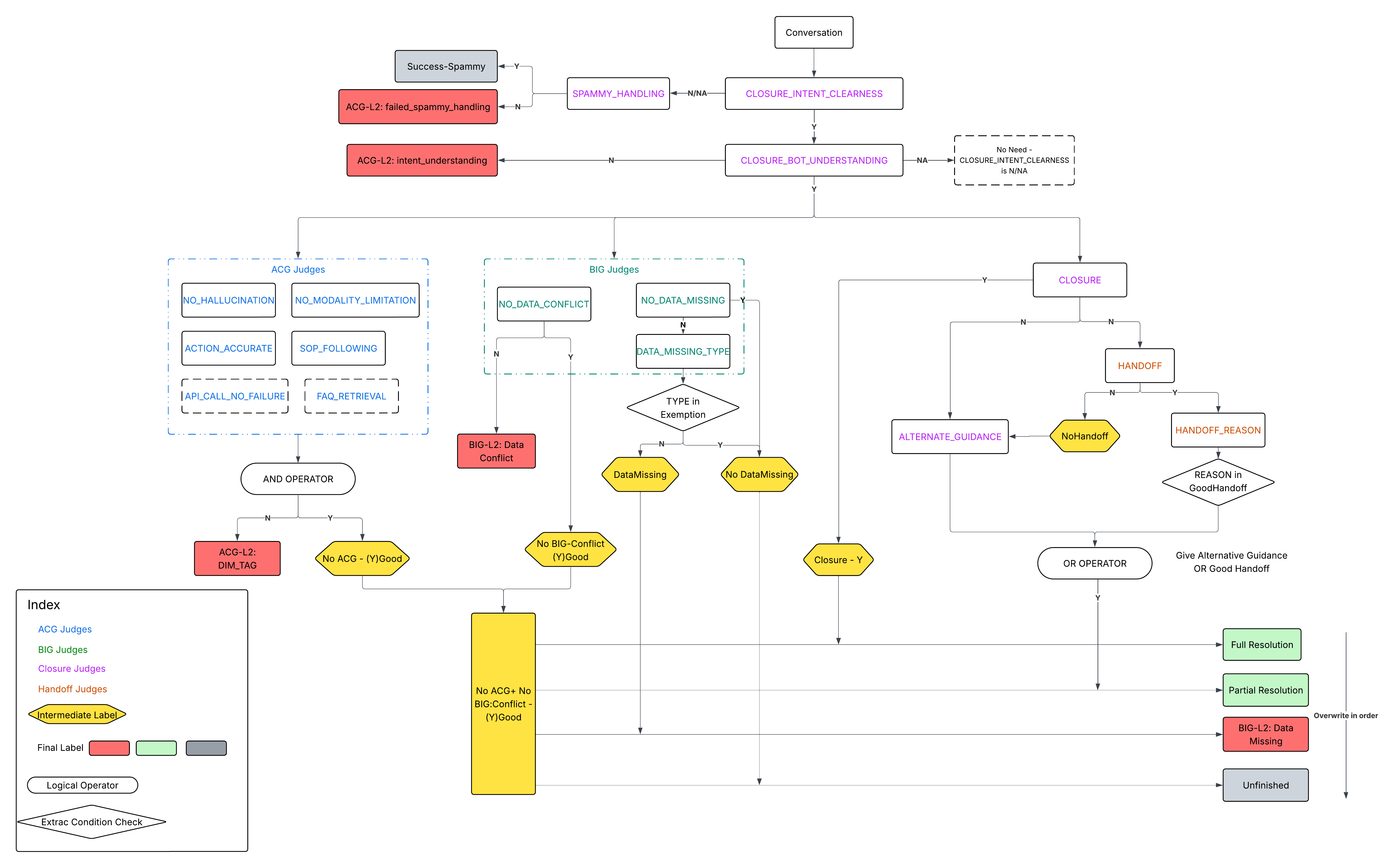}
\caption{The aggregation graph that implements the conversation-level specification. ACG denotes an agent-capability gap and BIG a business-information gap. The displayed Y/N values are raw predicate outcomes; downstream metrics use the metric-specific positive class defined above. Dashed nodes denote optional components outside the evaluated configuration.}
\label{fig:aggregation-logic}
\end{figure}

\paragraph{Label polarity is normalized before evaluation.}
The affirmative predicates shown in Figure~\ref{fig:aggregation-logic}, such as \texttt{NO\_HALLUCINATION} and \texttt{ACTION\_ACCURATE}, emit \texttt{Y} when the criterion is satisfied, so their raw \texttt{Y} means good. Defect-oriented metrics invert that representation so that positive means the target defect is present; success-oriented metrics retain successful completion or resolution as positive. Reported precision, recall, and NDR therefore refer to the named metric's target class, not directly to the raw \texttt{Y}/\texttt{N} token.

As Figure~\ref{fig:aggregation-logic} shows, the aggregation graph first gates on whether the user's intent is clear, whether the agent understood it, and whether an invalid or spammy conversation was handled correctly. For valid conversations, three branches run in parallel. \textbf{Agent-capability judges} cover hallucination, modality limitations, action accuracy, procedure following, and tool or retrieval failures. \textbf{Business-information judges} distinguish conflicting data from missing data and apply exemptions. \textbf{Outcome judges} evaluate task closure, handoff quality, and alternative guidance.

The aggregation layer then applies Boolean operators and explicit precedence rules to map overlapping atomic evidence into one conversation-level outcome. This separation is architectural: atomic judges can be developed and reused independently, while metric definitions change in the aggregation graph. It also makes the deployed metric version-specific---a change to a judge, operator, or overwrite order can change the reported result even when the underlying conversations remain fixed.

Table~\ref{tab:rca} summarises representative atomic judges and composite nodes in the aggregation graph.
It documents the implemented measurement structure; quantitative fidelity is reported separately in
Section~\ref{sec:validation}.

\begin{table}[t]
\centering
\caption{Structure of atomic and composite nodes in the aggregation graph.}
\label{tab:rca}
\scriptsize
\setlength{\tabcolsep}{3pt}
\begin{tabular}{@{}L{1.4cm} L{5.0cm} L{6.5cm}@{}}
\toprule
Level & Dimension & Aggregation role \\
\midrule
L5 & factual hallucination & atomic \\
L5 & procedure following & atomic \\
L4 & action accuracy & atomic \\
L3 & hallucination & \textbf{OR} of L4 + L5 \\
L3 & modality limitation & atomic \\
L3 & missing data & atomic \\
L3 & data conflict & atomic \\
L3 & intent understanding & atomic \\
L3 & spam handling & atomic \\
--- & task closure & atomic \\
--- & intent clearness & atomic \\
--- & alternative guidance & atomic \\
--- & good handoff & atomic \\
\textbf{aggregate} & \textbf{model-capability gap} & \textbf{OR} over L3--L5 \\
\textbf{aggregate} & knowledge-coverage gap & OR over business-information dimensions \\
L2 & full resolution & AND with negations \\
L2 & partial resolution & AND with negations \\
L2 & invalid conversation handled & AND with negations \\
L2 & unfinished conversation handled & AND with negations \\
\textbf{L2} & \textbf{model-capability gap only} & \textbf{AND with negation} \\
L2 & knowledge-coverage gap only & AND with negation \\
L2 & both gaps & AND with negation \\
L1 & failure & OR over all failure states \\
L1 & success & complement of failure \\
\bottomrule
\end{tabular}
\end{table}

\section{Simulation with an LLM-as-a-user}
\label{sec:simulation}

Pre-deployment evaluation of a multi-turn agent requires a user who can react when the agent under test deviates from a historical conversation. Human role-play is slow, inconsistent across annotators, and operationally expensive. Exact replay is deterministic, but a fixed response such as ``yes,'' ``that one,'' or ``how?'' can become incoherent when the agent asks a different question or changes the order of its answers. We therefore use an \textbf{LLM-as-a-user (LAAU)} to generate the user side of the interaction. \textbf{Design principle: specify and validate the simulated user as a measurement component before using its conversations to evaluate the agent.}

\subsection{Two LAAU operating settings}
\label{sec:7-1}

The same user model supports two complementary evaluation settings:

\begin{center}
\centering
\footnotesize
\renewcommand{\arraystretch}{1.18}
\setlength{\tabcolsep}{3pt}
\begin{tabular}{@{}L{0.18\textwidth} L{0.27\textwidth} L{0.25\textwidth} L{0.24\textwidth}@{}}
\toprule
Setting & Conditioning input & Required invariant & Primary uses \\
\midrule
\textbf{Conversation-conditioned LAAU} & Intents, answers, tone, and other user attributes extracted from an original conversation & Cover every original intent without introducing a new one & Regression and integration benchmarks; replaying failures after a bug fix \\
\textbf{Instruction-conditioned LAAU} & Predefined tasks, user traits, and optional user-profile facts & Enact the specified task and persona without exceeding the supplied information & Pre-onboarding tests; robustness, adversarial, JTBD, feature, and unit tests \\
\bottomrule
\end{tabular}
\end{center}

Conversation conditioning preserves the coverage of traffic-derived benchmarks while allowing the user to respond to a changed agent trajectory. Instruction conditioning creates cases that may be absent from history, including spam, penetration attempts, unclear or unusual intents, spelling variation, and anxious or otherwise atypical interaction styles. In both settings, the simulator produces only the user utterances visible to the agent and ends the interaction after the required intents have been exercised.

\subsection{Intent-preserving simulator design}
\label{sec:7-2}

The central invariant for conversation-conditioned LAAU is: \textbf{all original user intents must be expressed, and no new user intent may be introduced.} Persona, tone, wording, and turn order are secondary fidelity targets. This ordering matters because a fluent simulator that silently changes the task can create a false agent regression or improvement.

The implementation separates curation from generation. During benchmark construction, the system extracts the original user's intents and the answers they supplied to agent questions; audited benchmarks can instead provide human-verified key asks. During simulation, LAAU:

\begin{enumerate}[leftmargin=1.6em, itemsep=0.3ex, topsep=0.4ex]
\item answers an agent question only when the answer exists in the curated user information;
\item presents the remaining intents one at a time;
\item follows the original tone and phrasing where the current context still supports them;
\item minimally rephrases context-dependent messages when the agent trajectory has changed; and
\item terminates after all required intents have been addressed.
\end{enumerate}

The simulator sees the agent's final user-visible utterance, not its hidden instructions, retrievals, or reasoning. A versioned \textbf{user profile or user FAQ} supplies facts needed for authentication and replaces redacted values with synthetic but usable answers. Unknown answers remain unknown: unless a test explicitly specifies an uncooperative persona, the simulator does not invent information merely to keep the dialogue moving. The result is an adaptive replay rather than unrestricted role-play.

\subsection{Simulator fidelity specification}
\label{sec:7-3}

LAAU is evaluated independently of the agent judges using four criteria:

\begin{itemize}[leftmargin=1.6em, itemsep=0.3ex, topsep=0.4ex]
\item \textbf{Intent consistency:} all required intents are expressed and no additional intent is created.
\item \textbf{Persona consistency:} questions, answers, tone, emotion, and user-profile behaviour remain consistent with the source conversation or specified persona.
\item \textbf{Comprehensibility:} generated messages are understandable and do not leak internal instructions or model reasoning.
\item \textbf{Measurement stability:} repeated runs with an unchanged agent preserve the user task and should not materially move downstream resolution, model-capability, or knowledge-coverage metrics.
\end{itemize}

Human preference between LAAU and exact replay is retained as a diagnostic for conversational naturalness, but it is not a substitute for these fidelity criteria. A more natural conversation can still be invalid if it omits an intent or supplies information that the original user never provided.

\subsection{Validation requirements and limitations}
\label{sec:7-4}

Simulator release requires audited evidence for intent preservation, comprehensibility, run-to-run
stability, and downstream metric stability. Human preference between LAAU and exact replay is diagnostic,
but cannot establish task validity by itself. The limitations of the available validation evidence are
discussed in Section~\ref{sec:discussion-conclusion}.

Even after validation, LAAU is not assumed to be equivalent to real users. Minimal rephrasing can alter
tone; ambiguous, trolling, or intentionally incomplete messages invite subjective interpretation; and
different agent handoff points can move downstream metrics without a changed simulated intent. The
methodological claim is narrower: an LLM user becomes usable for multi-turn measurement when its intended
information, output boundary, fidelity criteria, and residual failure modes are explicitly specified and
audited.

\section{Human-in-the-loop evaluation and improvement}
\label{sec:human}

\begin{figure}[t]
\centering
\includegraphics[width=\linewidth]{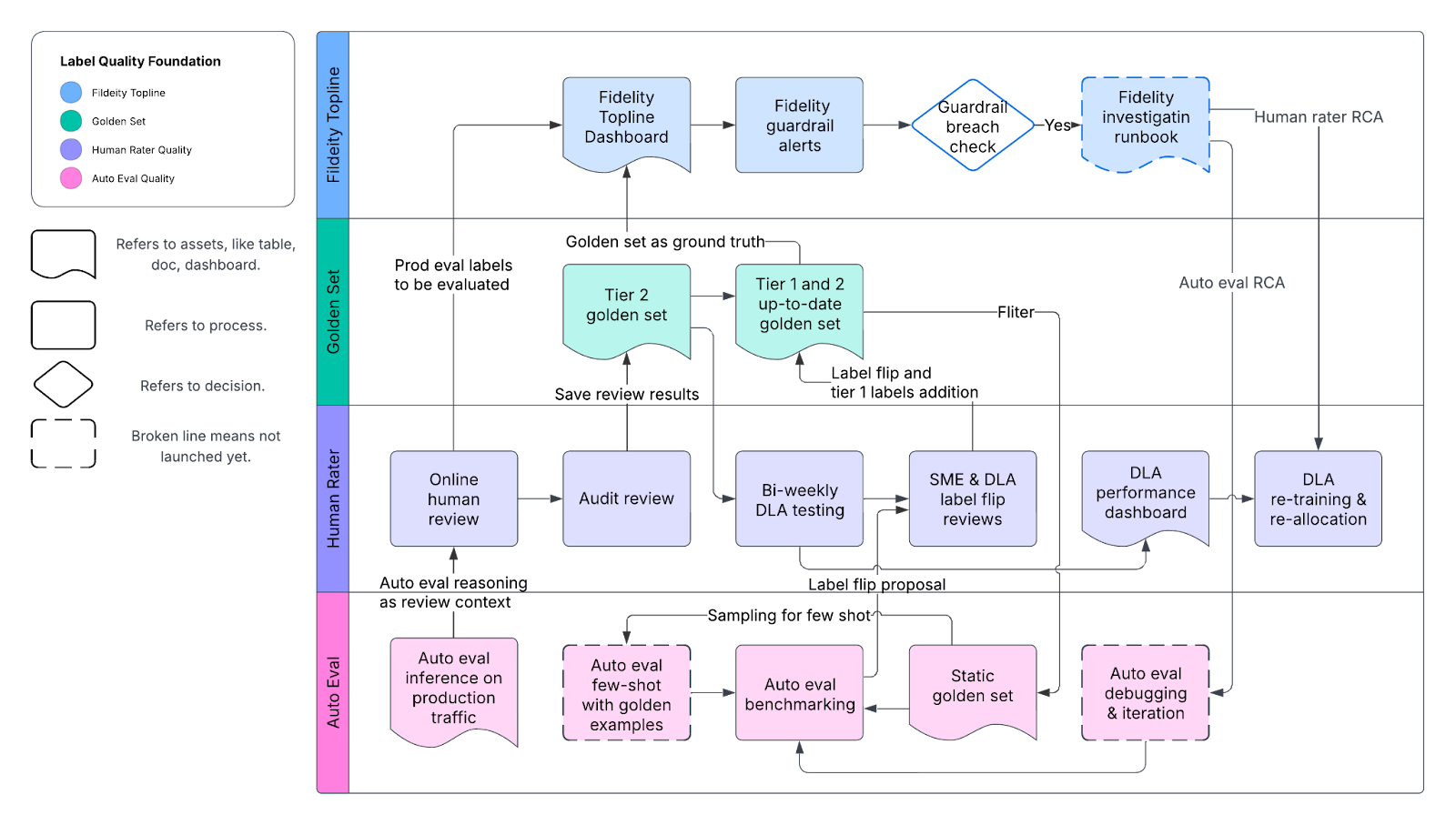}
\caption{The operational feedback system. Live judge outputs enter independent human review; audit results renew tiered golden sets; the golden sets test and allocate raters and benchmark judge revisions; fidelity dashboards and guardrails trigger investigation when the measurement system drifts. Dashed elements denote planned extensions.}
\label{fig:hitl-foundation}
\end{figure}

Human feedback maintains the measurement system after deployment by identifying drift and updating reference labels, specifications, and judges. The deployed loop is a label-quality control system with four connected layers: fidelity monitoring, tiered golden-set renewal, human-rater quality management, and judge benchmarking. A fidelity breach can therefore trigger both human-rater and judge root-cause analysis rather than being assigned to either side in advance. \textbf{Design principle: evaluate and improve the combined human--judge system, not either labeler in isolation.} Figure~\ref{fig:hitl-foundation} summarizes this feedback loop.

The judge-reasoning path in Figure~\ref{fig:hitl-foundation} applies only during escalation review; the initial human label remains independent of the judge output.

\subsection{Independent review and escalation}
\label{sec:human-review}

For each sampled conversation, the judge and an initial human reviewer label independently. Agreement produces the final label; disagreement routes the item to a quality-stratified escalation reviewer, who sees both labels and the judge's reasoning before making the final decision. A sample of these decisions is audited by top annotators, with a smaller expert-reviewed tier able to override them. Keeping the initial review blind mitigates anchoring and confirmation bias, while concentrating stronger reviewers on disputed items.

The measured outcome of this protocol is reported with the other real-world results in Section~\ref{sec:outcomes-alignment}.

\subsection{Reference maintenance and feedback}
\label{sec:human-maintenance}

Adjudicated disagreements can reveal a label error, specification ambiguity, or judge error. The workflow therefore routes them to rater correction, guideline revision, or judge improvement rather than assuming that either side is correct. This extends established work on annotation-error detection \citep{northcutt2021pervasive,chong2022detecting,nahum2024betterthanreported,chochlakis2025humanshallucinate} with an explicit specification-revision path.

As Figure~\ref{fig:hitl-foundation} shows, human review also maintains the reference system. Judges can produce the topline while humans audit for calibration and drift; audit volume is determined by the fidelity evidence required, not treated as a residual cost. Adjudicated examples renew a tiered golden set used to test and allocate raters, while unannounced golden jobs measure rater fidelity in the production queue. A separate static, expert-aligned set supports judge-release regression testing.

Golden-set quality bounds judge fidelity, and a finite release set cannot guarantee performance on changing live traffic. Reference labels must therefore be versioned and re-audited. The observed effects of the combined maintenance loop are reported in Section~\ref{sec:outcomes-alignment}.

\section{Empirical Studies}
\label{sec:validation}

We present two empirical studies of the methodology in production. The first examines whether system-level measurement fidelity improves over repeated judge iterations (Section~\ref{sec:outcomes-fidelity}). The second compares judge-only, human-only, and combined review workflows to characterize human--judge alignment and complementarity (Section~\ref{sec:outcomes-alignment}). The studies draw on production-derived data from multiple agents over two and a half years.

The studies answer different measurement questions and are observational rather than a controlled end-to-end comparison against an alternative methodology. Judge outputs were repeatedly audited against independently reviewed human labels on live traffic.

\subsection{Improvement in system-level measurement fidelity}
\label{sec:outcomes-fidelity}

\begin{center}
\small
\setlength{\tabcolsep}{4pt}
\begin{tabular}{@{}lll@{}}
\toprule
Audit stage & model-capability NDR & knowledge-coverage NDR \\
\midrule
Initial audit & \textbf{$-$0.022} & 0.299 \\
First follow-up & 0.22 & 0.45 \\
Second follow-up & 0.33 & 0.518 \\
Final reported audit & \textbf{0.550} & \textbf{0.686} \\
\emph{human co-pilot, final audit, same criteria} & \emph{0.611} & \emph{0.553} \\
\bottomrule
\end{tabular}
\end{center}

Three observations, each an argument for a specific part of the methodology.

\paragraph{The starting point was zero, and only the loop-level metric revealed it.}
A judge panel that had been in production for over a year, and was already good enough to be a useful co-pilot, had an NDR of \textbf{$-$0.022} the first time anyone measured it against an online auditor ground truth --- not detecting model-capability gaps better than chance. Every component-level metric looked acceptable; the co-pilot process had been carrying it. This is the strongest argument for a system-level fidelity scalar that no component can flatter.

\paragraph{The human comparator is imperfect and dimension-specific.}
An intermediate planning estimate put human NDR near 0.60; the final measurement put the human co-pilot at 0.611 on model capability and 0.553 on knowledge coverage. These are observed comparator values, not theoretical ceilings, and the paper lacks uncertainty intervals for the difference between human and judge readings.

\paragraph{The judge exceeded the observed human comparator on one dimension.}
On knowledge coverage the judge reached 0.686 against the human co-pilot's 0.553 at the final snapshot. The narrow operational reading is that systematic checking of machine-readable business context may suit automation; the snapshot does not establish general superiority over people. On model capability, the same period's summary was that the judge had better recall but worse precision.

\subsection{Alignment and complementarity with human raters}
\label{sec:outcomes-alignment}

\begin{table}[t]
\centering
\caption{Task-completion fidelity of human, judge, and combined review workflows. Precision and recall treat task completion as the positive class. \emph{Takeaway: the combined workflow has the strongest descriptive point estimate in both reported task-completion settings. No confidence intervals or paired significance tests were available, so the table does not establish statistical superiority.}}
\label{tab:copilot}
\footnotesize
\setlength{\tabcolsep}{3pt}
\begin{tabular}{@{}L{1.5cm} L{2.15cm} L{2.15cm} L{2.15cm} L{2.25cm} L{2.25cm}@{}}
\toprule
Agent & judge only P / R / acc & human only P / R / acc & \textbf{co-pilot P / R / acc} & \textbf{co-pilot $-$ judge $\Delta$ P / R / acc (pp)} & \textbf{co-pilot $-$ human $\Delta$ P / R / acc (pp)} \\
\midrule
Agent-A & 66.7 / 53.8 / 75.6 & 51.3 / 73.1 / 67.9 & \textbf{71.4 / 76.9 / 82.1} & \textbf{+4.7 / +23.1 / +6.5} & \textbf{+20.1 / +3.8 / +14.2} \\
Agent-B & 55.8 / 68.6 / 75.0 & 56.1 / 51.4 / 75.0 & \textbf{67.6 / 71.4 / 81.7} & \textbf{+11.8 / +2.8 / +6.7} & \textbf{+11.5 / +20.0 / +6.7} \\
\bottomrule
\end{tabular}
\end{table}

\begin{figure}[t]
\centering
\includegraphics[width=\linewidth]{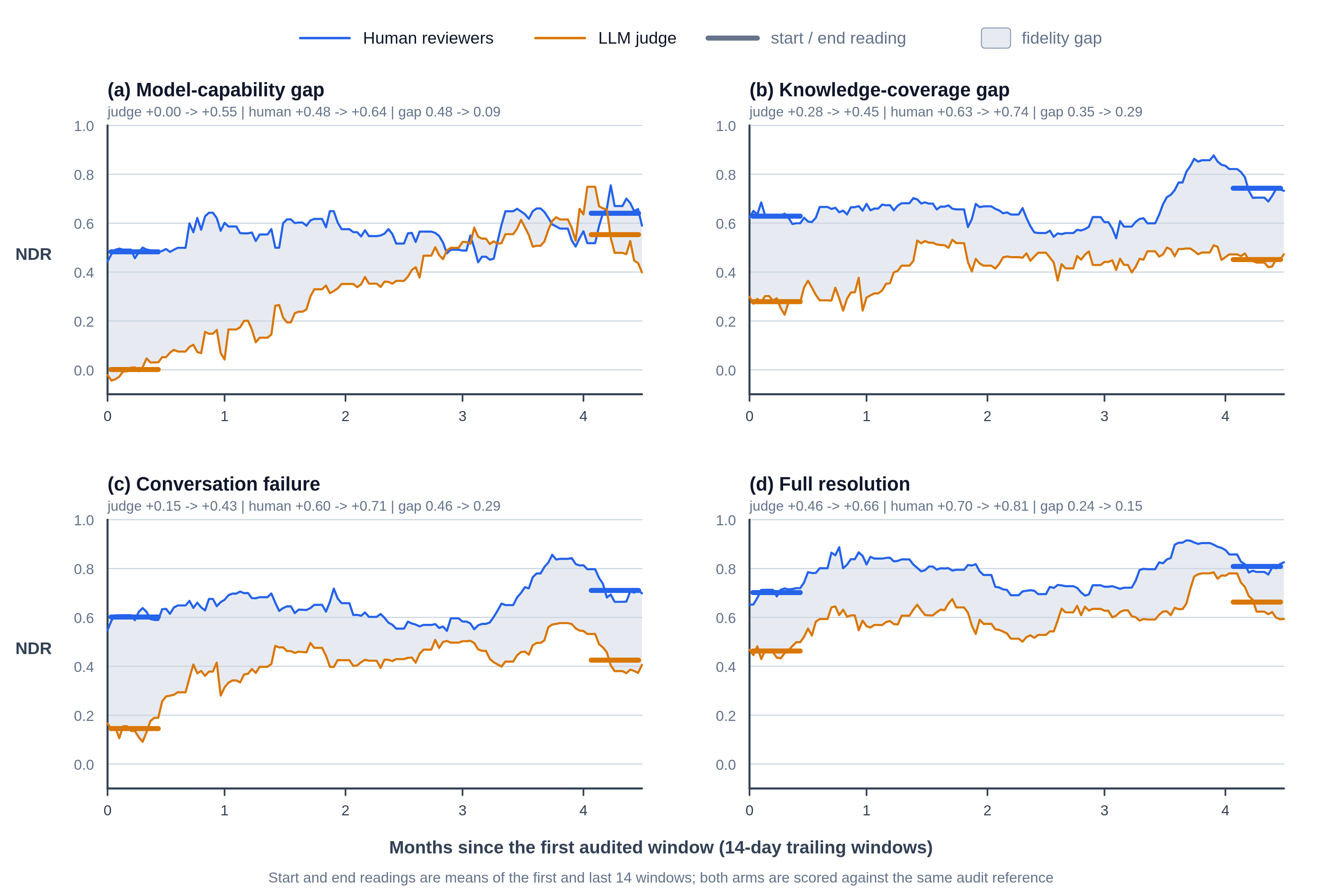}
\caption{Coupled improvement of human reviewers and the LLM judge, measured against the same audit reference. Each panel is one evaluation dimension for Agent-A in a single locale; both curves are net detection rate (TPR $-$ FPR) on 14-day trailing windows of live traffic, over the first 138 daily readings of the audited period. Time is given as months elapsed since the first audited window. Heavy segments mark the mean of the first and last fourteen windows --- the readings quoted in each panel subtitle --- and the shaded band is the human--judge fidelity gap. All eight series end above where they start, and the gap narrows on every dimension: model capability 0.48 to 0.09, conversation failure 0.46 to 0.29, knowledge coverage 0.35 to 0.29, full resolution 0.24 to 0.15. The audit continues past the plotted interval; Section~\ref{sec:discussion-conclusion} records what the later readings do. These NDR readings are not comparable with the precision and recall in Table~\ref{tab:copilot}, and a single continuous window from one agent and locale is observational: it shows co-movement under a shared feedback loop, not a causal effect of either arm on the other.}
\label{fig:coupled}
\end{figure}

Figure~\ref{fig:coupled} complements Table~\ref{tab:copilot} by tracking human-reviewer and judge fidelity over the same continuous audit window.

Across the continuous measurement in Figure~\ref{fig:coupled}, the judge's net detection rate rises on all four dimensions over four and a half months --- by \textbf{+0.55} on model capability, \textbf{+0.28} on conversation failure, \textbf{+0.20} on full resolution and \textbf{+0.17} on knowledge coverage --- starting from the same near-zero model-capability reading reported as the initial audit in Section~\ref{sec:outcomes-fidelity}. The human reviewers, scored by the same auditors against the same labels, rise by \textbf{+0.11 to +0.16} over the same window. Nothing in the series separates the two effects: guideline revisions, judge releases, rater training and audit-sample changes all fall inside the window, and the two arms share the specification, the disagreement review and the golden labels by construction (Sections~\ref{sec:reference-fidelity} and~\ref{sec:human-maintenance}). The claim the figure supports is therefore the weaker but still load-bearing one --- that a measurement system whose human and automated halves are coupled through one audited loop improved on both halves at once, rather than trading one against the other.

The gap narrows least on knowledge coverage, which is also the dimension on which the audit snapshots in Section~\ref{sec:outcomes-fidelity} put the judge \emph{above} the human comparator. The two are not the same measurement: that section reports discrete audit batches and Figure~\ref{fig:coupled} reports rolling windows for one agent and locale, and only their initial readings coincide. Taken together they indicate that the ordering between human and judge is dimension-specific and unstable across samples and across time, which is an argument for maintaining both arms and the audit that calibrates them, not for retiring either.

The observed pattern is consistent with complementarity rather than replacement. Relative to the judge, the co-pilot's advantage is concentrated in recall; relative to the human, it is concentrated in precision. The 2024 proof of concept likewise found that the judge did \textbf{not} outperform highly trained annotators on golden-set F1, while the combined workflow had the highest F1. The relevant success criterion is therefore alignment and error correction, not universal machine superiority.

The same loop improved the human reference process itself. Biweekly golden-set testing drove more than 70 reviewer reallocations, while breach investigations recovered L1-failure recall from \textbf{0.58 to 0.80} on one product and knowledge-gap recall from \textbf{0.24 to 0.81} on another. These are incident recoveries rather than controlled treatment effects, but they show that the feedback channel operated on human labels and workflow allocation as well as judge prompts.

\section{Discussion \& Conclusion}
\label{sec:discussion-conclusion}

These are observational production studies, not a controlled comparison against an alternative methodology. They support the operational usefulness of the complete methodology but do not isolate the causal effect of any individual architectural or workflow choice.

Figure~\ref{fig:coupled} plots the improving segment of a longer record, and the reader should treat it that way. The audit ran for a further ten weeks past the plotted interval. Over those weeks both arms are flat --- every reading moves by at most 0.05 except the judge's knowledge-coverage reading, which retraces by \textbf{0.07} --- giving back about two fifths of its gain over the plotted window. Auditing then stops being informative about the human arm altogether: its online labels stop disagreeing with the audit reference on every dimension, which makes the human reading unmeasurable rather than perfect, and no NDR after that point should be reported for it. The honest summary of the full record is therefore a period of coupled improvement followed by a plateau, not a sustained trend.

The human reference is also fallible. Expert-adjudicated labels were revised during disagreement review, the human comparator values in Section~\ref{sec:outcomes-fidelity} are observed comparators rather than ceilings, and the human--judge and co-pilot differences lack confidence intervals or paired significance tests. Precision/recall on static golden sets and NDR on live-traffic audits are therefore reported separately rather than chained.

Generalisability remains limited to one organisation and the reported agent, task, and locale samples. The LAAU evidence is limited to internal benchmarks and traffic-derived samples and does not establish equivalence to real human behaviour. These constraints bound the claim: the evidence does not show that the methodology is universally superior to human evaluation or to every alternative system design.

Reliable evaluation of multi-turn business agents requires more than a judge prompt. We presented a measurement-system methodology that connects a conversation-level specification, business-grounded atomic judges and explicit aggregation, intent-preserving user simulation, and an independently audited human feedback loop. The shared specification and reference process make failures actionable while allowing labels, guidelines, and judges to improve together.

The production studies show that system-level NDR improved across repeated audits, that the combined human--judge workflow had the strongest descriptive fidelity in both reported task-completion settings, and that the same feedback loop improved both automated and human evaluation. These observational results do not establish causal or universal superiority, but they support treating LLM-based evaluation as a maintained measurement system whose fidelity must be continuously demonstrated as the agent, evidence, and traffic evolve.


\bibliographystyle{plainnat}
\bibliography{paper}

@misc{zheng2023judging,
  title={{Judging LLM-as-a-Judge with MT-Bench and Chatbot Arena}},
  author={Lianmin Zheng and Wei-Lin Chiang and Ying Sheng and Siyuan Zhuang and Zhanghao Wu and Yonghao Zhuang and Zi Lin and Zhuohan Li and Dacheng Li and Eric P. Xing and Hao Zhang and Joseph E. Gonzalez and Ion Stoica},
  year={2023}, eprint={2306.05685}, archivePrefix={arXiv},
  note={NeurIPS 2023 Datasets and Benchmarks Track},
  url={https://arxiv.org/abs/2306.05685}}

@misc{liu2023geval,
  title={{G-Eval: NLG Evaluation using GPT-4 with Better Human Alignment}},
  author={Yang Liu and Dan Iter and Yichong Xu and Shuohang Wang and Ruochen Xu and Chenguang Zhu},
  year={2023}, eprint={2303.16634}, archivePrefix={arXiv},
  url={https://arxiv.org/abs/2303.16634}}

@misc{kim2023prometheus,
  title={{Prometheus: Inducing Fine-grained Evaluation Capability in Language Models}},
  author={Seungone Kim and Jamin Shin and Yejin Cho and Joel Jang and Shayne Longpre and Hwaran Lee and Sangdoo Yun and Seongjin Shin and Sungdong Kim and James Thorne and Minjoon Seo},
  year={2023}, eprint={2310.08491}, archivePrefix={arXiv},
  note={ICLR 2024},
  url={https://arxiv.org/abs/2310.08491}}

@misc{li2024generation,
  title={{From Generation to Judgment: Opportunities and Challenges of LLM-as-a-judge}},
  author={Dawei Li and Bohan Jiang and Liangjie Huang and Alimohammad Beigi and Chengshuai Zhao and Zhen Tan and Amrita Bhattacharjee and Yuxuan Jiang and Canyu Chen and Tianhao Wu and Kai Shu and Lu Cheng and Huan Liu},
  year={2024}, eprint={2411.16594}, archivePrefix={arXiv},
  note={EMNLP 2025},
  url={https://arxiv.org/abs/2411.16594}}

@misc{chiang2023can,
  title={{Can Large Language Models Be an Alternative to Human Evaluations?}},
  author={Cheng-Han Chiang and Hung-yi Lee},
  year={2023}, eprint={2305.01937}, archivePrefix={arXiv},
  note={ACL 2023 main conference},
  url={https://arxiv.org/abs/2305.01937}}

@misc{shi2024judgingposition,
  title={{Judging the Judges: A Systematic Study of Position Bias in LLM-as-a-Judge}},
  author={Lin Shi and Chiyu Ma and Wenhua Liang and Xingjian Diao and Weicheng Ma and Soroush Vosoughi},
  year={2024}, eprint={2406.07791}, archivePrefix={arXiv},
  note={AACL-IJCNLP 2025},
  url={https://arxiv.org/abs/2406.07791}}

@misc{lee2025correctly,
  title={{How to Correctly Report LLM-as-a-Judge Evaluations}},
  author={Chungpa Lee and Thomas Zeng and Jongwon Jeong and Jy-yong Sohn and Kangwook Lee},
  year={2025}, eprint={2511.21140}, archivePrefix={arXiv},
  note={International Conference on Machine Learning (ICML) 2026},
  url={https://arxiv.org/abs/2511.21140}}

@misc{tan2024judgebench,
  title={{JudgeBench: A Benchmark for Evaluating LLM-based Judges}},
  author={Sijun Tan and Siyuan Zhuang and Kyle Montgomery and William Y. Tang and Alejandro Cuadron and Chenguang Wang and Raluca Ada Popa and Ion Stoica},
  year={2024}, eprint={2410.12784}, archivePrefix={arXiv},
  note={ICLR 2025},
  url={https://arxiv.org/abs/2410.12784}}

@misc{thakur2024judgingjudges,
  title={{Judging the Judges: Evaluating Alignment and Vulnerabilities in LLMs-as-Judges}},
  author={Aman Singh Thakur and Kartik Choudhary and Venkat Srinik Ramayapally and Sankaran Vaidyanathan and Dieuwke Hupkes},
  year={2024}, eprint={2406.12624}, archivePrefix={arXiv},
  note={Proceedings of the Fourth Workshop on Generation, Evaluation and Metrics (GEM2) 2025, pp. 404-430},
  url={https://arxiv.org/abs/2406.12624}}

@misc{guan2025multiturnsurvey,
  title={{Evaluating LLM-based Agents for Multi-Turn Conversations: A Survey}},
  author={Shengyue Guan and Jindong Wang and Jiang Bian and Bin Zhu and Jian-guang Lou and Haoyi Xiong},
  year={2025}, eprint={2503.22458}, archivePrefix={arXiv},
  url={https://arxiv.org/abs/2503.22458}}

@misc{kwan2024mteval,
  title={{MT-Eval: A Multi-Turn Capabilities Evaluation Benchmark for Large Language Models}},
  author={Wai-Chung Kwan and Xingshan Zeng and Yuxin Jiang and Yufei Wang and Liangyou Li and Lifeng Shang and Xin Jiang and Qun Liu and Kam-Fai Wong},
  year={2024}, eprint={2401.16745}, archivePrefix={arXiv},
  url={https://arxiv.org/abs/2401.16745}}

@misc{bai2024mtbench101,
  title={{MT-Bench-101: A Fine-Grained Benchmark for Evaluating Large Language Models in Multi-Turn Dialogues}},
  author={Ge Bai and Jie Liu and Xingyuan Bu and Yancheng He and Jiaheng Liu and Zhanhui Zhou and Zhuoran Lin and Wenbo Su and Tiezheng Ge and Bo Zheng and Wanli Ouyang},
  year={2024}, eprint={2402.14762}, archivePrefix={arXiv},
  note={ACL 2024},
  url={https://arxiv.org/abs/2402.14762}}

@misc{wang2023mint,
  title={{MINT: Evaluating LLMs in Multi-turn Interaction with Tools and Language Feedback}},
  author={Xingyao Wang and Zihan Wang and Jiateng Liu and Yangyi Chen and Lifan Yuan and Hao Peng and Heng Ji},
  year={2023}, eprint={2309.10691}, archivePrefix={arXiv},
  note={ICLR 2024},
  url={https://arxiv.org/abs/2309.10691}}

@misc{yao2024taubench,
  title={{$\tau$-bench: A Benchmark for Tool-Agent-User Interaction in Real-World Domains}},
  author={Shunyu Yao and Noah Shinn and Pedram Razavi and Karthik Narasimhan},
  year={2024}, eprint={2406.12045}, archivePrefix={arXiv},
  url={https://arxiv.org/abs/2406.12045}}

@misc{barres2025tau2bench,
  title={{$\tau^2$-Bench: Evaluating Conversational Agents in a Dual-Control Environment}},
  author={Victor Barres and Honghua Dong and Soham Ray and Xujie Si and Karthik Narasimhan},
  year={2025}, eprint={2506.07982}, archivePrefix={arXiv},
  url={https://arxiv.org/abs/2506.07982}}

@misc{zhou2023sotopia,
  title={{SOTOPIA: Interactive Evaluation for Social Intelligence in Language Agents}},
  author={Xuhui Zhou and Hao Zhu and Leena Mathur and Ruohong Zhang and Haofei Yu and Zhengyang Qi and Louis-Philippe Morency and Yonatan Bisk and Daniel Fried and Graham Neubig and Maarten Sap},
  year={2023}, eprint={2310.11667}, archivePrefix={arXiv},
  url={https://arxiv.org/abs/2310.11667}}

@misc{arcadinho2024almita,
  title={{Automated test generation to evaluate tool-augmented LLMs as conversational AI agents}},
  author={Samuel Arcadinho and David Aparicio and Mariana Almeida},
  year={2024}, eprint={2409.15934}, archivePrefix={arXiv},
  note={GenBench@EMNLP 2024},
  url={https://arxiv.org/abs/2409.15934}}

@misc{huang2024crmarena,
  title={{CRMArena: Understanding the Capacity of LLM Agents to Perform Professional CRM Tasks in Realistic Environments}},
  author={Kung-Hsiang Huang and Akshara Prabhakar and Sidharth Dhawan and Yixin Mao and Huan Wang and Silvio Savarese and Caiming Xiong and Philippe Laban and Chien-Sheng Wu},
  year={2024}, eprint={2411.02305}, archivePrefix={arXiv},
  note={NAACL 2025},
  url={https://arxiv.org/abs/2411.02305}}

@misc{fan2026shoppingreasoning,
  title={{Shopping Reasoning Bench: An Expert-Authored Benchmark for Multi-Turn Conversational Shopping Assistants}},
  author={Shuxian Fan and Seonwoo Min and Youna Hu and Botao Xia and Jayakrishnan Unnikrishnan and Rowan Musselmann and Yifan Gao and Qingyu Yin and Priyanka Nigam and Bing Yin},
  year={2026}, eprint={2606.12608}, archivePrefix={arXiv},
  url={https://arxiv.org/abs/2606.12608}}

@misc{mehri2020fed,
  title={{Unsupervised Evaluation of Interactive Dialog with DialoGPT}},
  author={Shikib Mehri and Maxine Eskenazi},
  year={2020}, eprint={2006.12719}, archivePrefix={arXiv},
  note={SIGdial 2020},
  url={https://arxiv.org/abs/2006.12719}}

@misc{plank2022labelvariation,
  title={{The 'Problem' of Human Label Variation: On Ground Truth in Data, Modeling and Evaluation}},
  author={Barbara Plank},
  year={2022}, eprint={2211.02570}, archivePrefix={arXiv},
  note={EMNLP 2022},
  url={https://arxiv.org/abs/2211.02570}}

@misc{parmar2022dontblame,
  title={{Don't Blame the Annotator: Bias Already Starts in the Annotation Instructions}},
  author={Mihir Parmar and Swaroop Mishra and Mor Geva and Chitta Baral},
  year={2022}, eprint={2205.00415}, archivePrefix={arXiv},
  note={EACL 2023 (Outstanding Paper Award)},
  url={https://arxiv.org/abs/2205.00415}}

@misc{northcutt2021pervasive,
  title={{Pervasive Label Errors in Test Sets Destabilize Machine Learning Benchmarks}},
  author={Curtis G. Northcutt and Anish Athalye and Jonas Mueller},
  year={2021}, eprint={2103.14749}, archivePrefix={arXiv},
  note={NeurIPS 2021 Track on Datasets and Benchmarks},
  url={https://arxiv.org/abs/2103.14749}}

@misc{chong2022detecting,
  title={{Detecting Label Errors by using Pre-Trained Language Models}},
  author={Derek Chong and Jenny Hong and Christopher D. Manning},
  year={2022}, eprint={2205.12702}, archivePrefix={arXiv},
  note={EMNLP 2022},
  url={https://arxiv.org/abs/2205.12702}}

@misc{nahum2024betterthanreported,
  title={{Are LLMs Better than Reported? Detecting Label Errors and Mitigating Their Effect on Model Performance}},
  author={Omer Nahum and Nitay Calderon and Orgad Keller and Idan Szpektor and Roi Reichart},
  year={2024}, eprint={2410.18889}, archivePrefix={arXiv},
  url={https://arxiv.org/abs/2410.18889}}

@misc{chochlakis2025humanshallucinate,
  title={{Humans Hallucinate Too: Language Models Identify and Correct Subjective Annotation Errors With Label-in-a-Haystack Prompts}},
  author={Georgios Chochlakis and Peter Wu and Arjun Bedi and Marcus Ma and Kristina Lerman and Shrikanth Narayanan},
  year={2025}, eprint={2505.17222}, archivePrefix={arXiv},
  note={EMNLP 2025 Main Proceedings},
  url={https://arxiv.org/abs/2505.17222}}

@misc{gupta2026nubank,
  title={{Building Customer Support AI Agents at 100M-User Scale: An Evaluation-Driven Framework}},
  author={Aman Gupta and Kevin Rossell and Edesio Alcobaça and Jose Chrystian Lima Pacheco and Carolina Baptista de Lima and Shao Tang and Luiz Paulo Rabachini and Luis Moneda and Herbert Fei and Daniel Silva and Rohan Ramanath},
  year={2026}, eprint={2606.08867}, archivePrefix={arXiv},
  note={KDD '26 (32nd ACM SIGKDD Conference on Knowledge Discovery and Data Mining)},
  url={https://arxiv.org/abs/2606.08867}}

@misc{zhang2026catchingonefive,
  title={{Catching One in Five: LLM-as-Judge Blind Spots in Production Multi-Turn Transaction Agents}},
  author={Sawyer Zhang and Alexander Wang and Sophie Lei},
  year={2026}, eprint={2606.10315}, archivePrefix={arXiv},
  url={https://arxiv.org/abs/2606.10315}}

@misc{maharaj2024enterprise,
  title={{Evaluation and Continual Improvement for an Enterprise AI Assistant}},
  author={Akash V. Maharaj and Kun Qian and Uttaran Bhattacharya and Sally Fang and Horia Galatanu and Manas Garg and Rachel Hanessian and Nishant Kapoor and Ken Russell and Shivakumar Vaithyanathan and Yunyao Li},
  year={2024}, eprint={2407.12003}, archivePrefix={arXiv},
  note={DaSH Workshop at NAACL 2024},
  url={https://arxiv.org/abs/2407.12003}}

@misc{shankar2024validators,
  title={{Who Validates the Validators? Aligning LLM-Assisted Evaluation of LLM Outputs with Human Preferences}},
  author={Shreya Shankar and J. D. Zamfirescu-Pereira and Björn Hartmann and Aditya G. Parameswaran and Ian Arawjo},
  year={2024}, eprint={2404.12272}, archivePrefix={arXiv},
  url={https://arxiv.org/abs/2404.12272}}

@misc{markov2022moderation,
  title={{A Holistic Approach to Undesired Content Detection in the Real World}},
  author={Todor Markov and Chong Zhang and Sandhini Agarwal and Tyna Eloundou and Teddy Lee and Steven Adler and Angela Jiang and Lilian Weng},
  year={2022}, eprint={2208.03274}, archivePrefix={arXiv},
  note={Oral presentation at AAAI-23},
  url={https://arxiv.org/abs/2208.03274}}

@misc{xu2024ragkg,
  title={{Retrieval-Augmented Generation with Knowledge Graphs for Customer Service Question Answering}},
  author={Zhentao Xu and Mark Jerome Cruz and Matthew Guevara and Tie Wang and Manasi Deshpande and Xiaofeng Wang and Zheng Li},
  year={2024}, eprint={2404.17723}, archivePrefix={arXiv},
  doi={10.1145/3626772.3661370},
  url={https://arxiv.org/abs/2404.17723}}

@misc{ashktorab2025evalassist,
  title={{EvalAssist: A Human-Centered Tool for LLM-as-a-Judge}},
  author={Zahra Ashktorab and Werner Geyer and Michael Desmond and Elizabeth M. Daly and Martin Santillan Cooper and Qian Pan and Erik Miehling and Tejaswini Pedapati and Hyo Jin Do},
  year={2025}, eprint={2507.02186}, archivePrefix={arXiv},
  url={https://arxiv.org/abs/2507.02186}}

@misc{ye2023flask,
  title={{FLASK: Fine-grained Language Model Evaluation based on Alignment Skill Sets}},
  author={Seonghyeon Ye and Doyoung Kim and Sungdong Kim and Hyeonbin Hwang and Seungone Kim and Yongrae Jo and James Thorne and Juho Kim and Minjoon Seo},
  year={2023}, eprint={2307.10928}, archivePrefix={arXiv},
  note={ICLR 2024 Spotlight},
  url={https://arxiv.org/abs/2307.10928}}

@misc{cook2024ticking,
  title={{TICKing All the Boxes: Generated Checklists Improve LLM Evaluation and Generation}},
  author={Jonathan Cook and Tim Rocktäschel and Jakob Foerster and Dennis Aumiller and Alex Wang},
  year={2024}, eprint={2410.03608}, archivePrefix={arXiv},
  url={https://arxiv.org/abs/2410.03608}}

@misc{lee2024checkeval,
  title={{CheckEval: A reliable LLM-as-a-Judge framework for evaluating text generation using checklists}},
  author={Yukyung Lee and Joonghoon Kim and Jaehee Kim and Hyowon Cho and Jaewook Kang and Pilsung Kang and Najoung Kim},
  year={2024}, eprint={2403.18771}, archivePrefix={arXiv},
  note={EMNLP 2025},
  url={https://arxiv.org/abs/2403.18771}}

@misc{li2024dnaeval,
  title={{DnA-Eval: Enhancing Large Language Model Evaluation through Decomposition and Aggregation}},
  author={Minzhi Li and Zhengyuan Liu and Shumin Deng and Shafiq Joty and Nancy F. Chen and Min-Yen Kan},
  year={2024}, eprint={2405.15329}, archivePrefix={arXiv},
  note={COLING 2025},
  url={https://arxiv.org/abs/2405.15329}}

@misc{hashemi2024llmrubric,
  title={{LLM-Rubric: A Multidimensional, Calibrated Approach to Automated Evaluation of Natural Language Texts}},
  author={Helia Hashemi and Jason Eisner and Corby Rosset and Benjamin Van Durme and Chris Kedzie},
  year={2024}, eprint={2501.00274}, archivePrefix={arXiv},
  note={Proceedings of ACL 2024 (Volume 1: Long Papers), pp. 13806-13834},
  url={https://arxiv.org/abs/2501.00274}}

@misc{liu2024hdeval,
  title={{HD-Eval: Aligning Large Language Model Evaluators Through Hierarchical Criteria Decomposition}},
  author={Yuxuan Liu and Tianchi Yang and Shaohan Huang and Zihan Zhang and Haizhen Huang and Furu Wei and Weiwei Deng and Feng Sun and Qi Zhang},
  year={2024}, eprint={2402.15754}, archivePrefix={arXiv},
  url={https://arxiv.org/abs/2402.15754}}

@misc{wei2025rocketeval,
  title={{RocketEval: Efficient Automated LLM Evaluation via Grading Checklist}},
  author={Tianjun Wei and Wei Wen and Ruizhi Qiao and Xing Sun and Jianghong Ma},
  year={2025}, eprint={2503.05142}, archivePrefix={arXiv},
  note={ICLR 2025},
  url={https://arxiv.org/abs/2503.05142}}

@misc{arora2025healthbench,
  title={{HealthBench: Evaluating Large Language Models Towards Improved Human Health}},
  author={Rahul K. Arora and Jason Wei and Rebecca Soskin Hicks and Preston Bowman and Joaquin Quiñonero-Candela and Foivos Tsimpourlas and Michael Sharman and Meghan Shah and Andrea Vallone and Alex Beutel and Johannes Heidecke and Karan Singhal},
  year={2025}, eprint={2505.08775}, archivePrefix={arXiv},
  url={https://arxiv.org/abs/2505.08775}}

@misc{viswanathan2025checklists,
  title={{Checklists Are Better Than Reward Models For Aligning Language Models}},
  author={Vijay Viswanathan and Yanchao Sun and Shuang Ma and Xiang Kong and Meng Cao and Graham Neubig and Tongshuang Wu},
  year={2025}, eprint={2507.18624}, archivePrefix={arXiv},
  note={NeurIPS 2025},
  url={https://arxiv.org/abs/2507.18624}}

@misc{cho2026bineval,
  title={{Ask, Don't Judge: Binary Questions for Interpretable LLM Evaluation and Self-Improvement}},
  author={Sangwoo Cho and Kushal Chawla and Pengshan Cai and Zefang Liu and Chenyang Zhu and Shi-Xiong Zhang and Sambit Sahu},
  year={2026}, eprint={2606.27226}, archivePrefix={arXiv},
  note={Second Workshop on Compositional Learning at ICML 2026},
  url={https://arxiv.org/abs/2606.27226}}

@misc{zhang2026rethinkingatomic,
  title={{A Matched Holistic Rubric Rivals Self-Decomposing Atomic Judges for Benchmark-Style Reference-Support Classification}},
  author={Xinran Zhang},
  year={2026}, eprint={2603.28005}, archivePrefix={arXiv},
  url={https://arxiv.org/abs/2603.28005}}

@misc{rao2026autorubric,
  title={{Autorubric: A Unifying Framework for Rubric-Based LLM Evaluation on Non-Verifiable Tasks}},
  author={Delip Rao and Chris Callison-Burch},
  year={2026}, eprint={2603.00077}, archivePrefix={arXiv},
  note={COLM 2026},
  url={https://arxiv.org/abs/2603.00077}}

@article{youden1950index,
  title={{Index for Rating Diagnostic Tests}},
  author={Youden, W. J.},
  journal={Cancer},
  volume={3},
  number={1},
  pages={32--35},
  year={1950},
  doi={10.1002/1097-0142(1950)3:1<32::AID-CNCR2820030106>3.0.CO;2-3}}

\end{document}